\documentclass[10pt,twocolumn,letterpaper]{article}

\usepackage{config_template/iccv}      
\usepackage{times}
\usepackage{epsfig}
\usepackage{graphicx}
\usepackage{amsmath}
\usepackage{amssymb}

\definecolor{iccvblue}{rgb}{0.21,0.49,0.74}
\usepackage[pagebackref,breaklinks,colorlinks,allcolors=iccvblue]{hyperref}

\def\paperID{430} 
\def\confName{ICCV}
\def\confYear{2025}

\usepackage[accsupp]{axessibility}

\usepackage{lineno}
\usepackage{booktabs}

\usepackage[inline]{enumitem}

\usepackage[normalem]{ulem}

\usepackage{wrapfig}
\usepackage{subcaption}

\usepackage{url}

\usepackage{bm}
\usepackage{xspace}
\usepackage{caption}

\usepackage{balance}

\usepackage{array, makecell}

\usepackage{pifont}

\usepackage{multirow}

\usepackage[capitalize]{cleveref}
\crefname{section}{Sec.}{Secs.}
\Crefname{section}{Section}{Sections}
\Crefname{table}{Table}{Tables}
\crefname{table}{Tab.}{Tabs.}
\crefname{figure}{Fig.}{Figs.}
\Crefname{figure}{Figure}{Figures}

\usepackage{tabularx}

\newcommand{\ourmethod}{\mbox{\textcolor{\nameCOLOR}{BioTUCH}}\xspace}

\newcommand{\nameCOLOR}{black}

\newcommand{\tuch}{\mbox{TUCH}\xspace}
\newcommand{\multihmr}{\mbox{Multi-HMR}\xspace}
\newcommand{\aios}{\mbox{AiOS}\xspace}

\newcommand{\websiteURL}{\mbox{\href{https://biotuch.is.tue.mpg.de}{biotuch.is.tue.mpg.de}}}

\newcommand{\ourTitle}{Contact-Aware Refinement of Human Pose Pseudo-Ground Truth\\via Bioimpedance Sensing}

\newcommand{\smplx}{\mbox{SMPL-X}\xspace}

\newcommand{\smplifyx}{\mbox{SMPLify-X}\xspace}

\newcommand{\mocap}{\mbox{mocap}\xspace}

\newcommand{\threeD}{3D\xspace}

\newcommand{\etal}{\emph{et al}.\xspace}
\newcommand{\ie}{\emph{i.e}.,\xspace}
\newcommand{\eg}{\emph{e.g}.,\xspace}

\newcommand{\rgb}{\mbox{RGB}\xspace}

\newcommand{\qheading}[1]{\noindent\textbf{#1:}}

\newcommand{\PAvtov}{\mbox{PA-V2V}\xspace}

\newcommand{\st}{self-touch\xspace}

\newcommand{\sct}{self-contact\xspace}
\newcommand{\gt}{GT\xspace}
\newcommand{\pgt}{pGT\xspace}

\usepackage[dvipsnames]{xcolor}

\begin{document}

\title{\ourTitle} 

\author{%
Maria-Paola~Forte \quad
Nikos~Athanasiou\textsuperscript{*} \quad
Giulia~Ballardini\textsuperscript{*} \quad
Jan~Ulrich~Bartels \and
Katherine~J.~Kuchenbecker \quad
Michael~J.~Black\\
{\small Max Planck Institute for Intelligent Systems, Stuttgart and T{\"u}bingen, Germany}\\
\tt\small{\{forte,ballardini,jub,kjk\}@is.mpg.de} \quad \{nikos.athanasiou,black\}@tue.mpg.de
\vspace{-2.0em}
}
\twocolumn[{
    \renewcommand\twocolumn[1][]{#1}
    \maketitle
    \centering
        \begin{minipage}{\textwidth}
            \centering
            \includegraphics[width=0.67\textwidth,clip=true,trim=0.5cm 1.6cm 16.5cm 0cm]{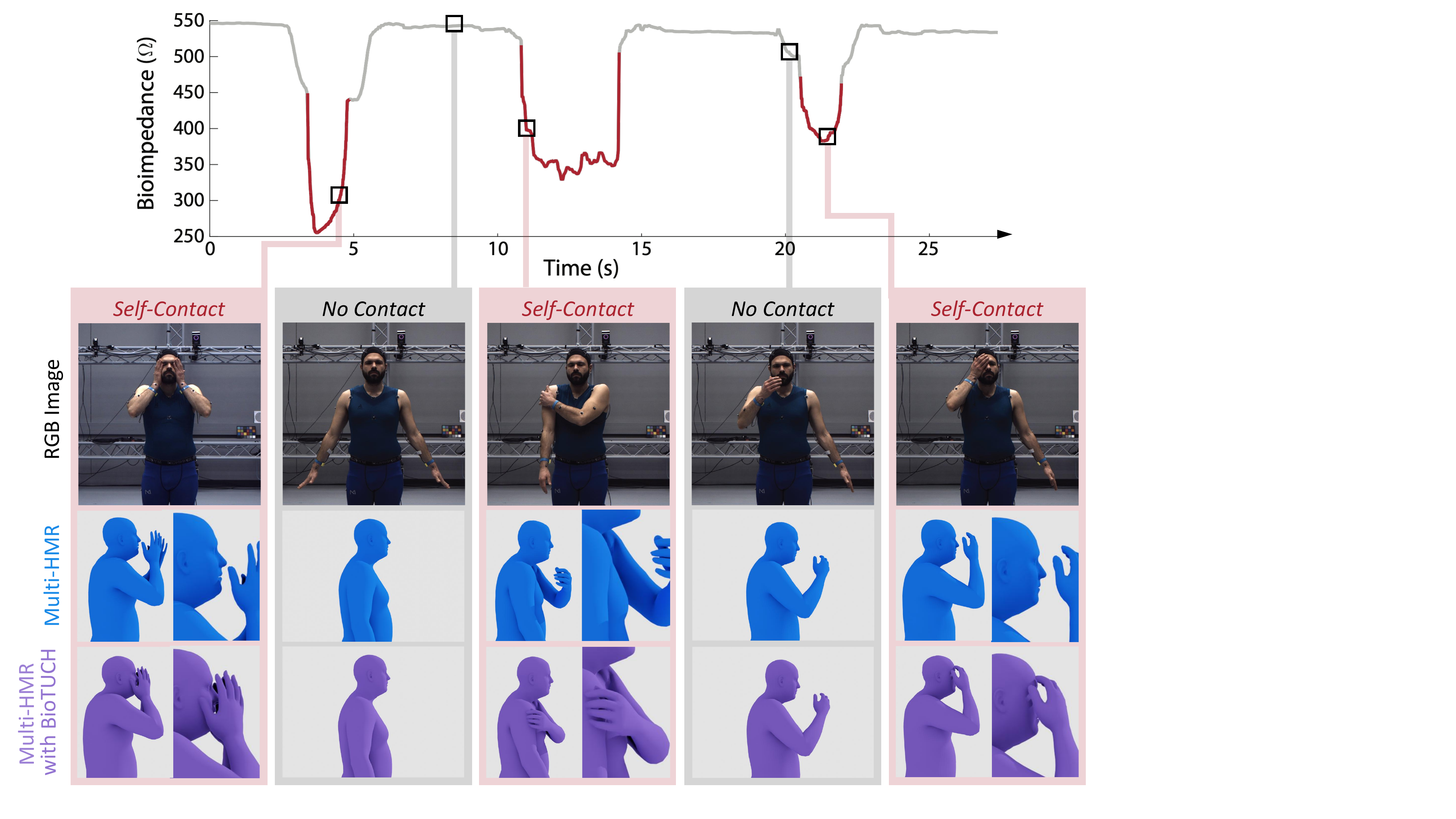}
        \end{minipage}
    \captionof{figure}{Human pose estimation methods struggle to reconstruct \sct along the camera's viewing axis. \ourmethod uses sharp changes in the bioimpedance signal measured between the wrists to estimate the beginning and end of such contacts. For all frames between these points (red segments of the signal), \ourmethod optimizes the results of off-the-shelf methods (such as \multihmr) to create plausible contact. Five selected poses (at the indicated time points) and their reconstructions are shown, with zoomed-in views of the contact regions.\looseness=-1}
    \label{fig:teaser}
    \vspace*{+01.5em}
}]

\maketitle
\renewcommand{\thefootnote}{\fnsymbol{footnote}}
\footnotetext[1]{Equal contribution.}
\renewcommand{\thefootnote}{\arabic{footnote}}
\begin{abstract}
Capturing accurate 3D human pose in the wild would provide valuable data for training pose estimation and motion generation methods.
While video-based estimation approaches have become increasingly accurate, they often fail in common scenarios involving \sct, such as a hand touching the face.
In contrast, wearable bioimpedance sensing can cheaply and unobtrusively measure ground-truth skin-to-skin contact.
Consequently, we propose a novel framework that combines visual pose estimators with bioimpedance sensing to capture the 3D pose of people by taking \sct into account.
Our method, \ourmethod, initializes the pose using an off-the-shelf estimator and introduces contact-aware pose optimization during measured \sct: reprojection error and deviations from the input estimate are minimized while enforcing vertex proximity constraints.
We validate our approach using a new dataset of synchronized RGB video, bioimpedance measurements, and 3D motion capture. Testing with three input pose estimators, we demonstrate an average of 11.7\% improvement in reconstruction accuracy.
We also present a miniature wearable bioimpedance sensor that enables efficient large-scale collection of contact-aware training data for improving pose estimation and generation using \ourmethod.
Code and data are available at \websiteURL 
\end{abstract}
\section{Introduction}

Self-touch is an inherent part of human motion, whether scratching an arm or using the hands to support the head.
It occurs frequently and plays a vital role in functional movements, psychological expression, comfort behaviors, and communicative acts.
Thus, understanding and accurately modeling \st (defined as hand-initiated contact, \eg a hand touching the face, clasped hands) and more generally \sct (contact between any body parts, \eg ankle-to-ankle) would greatly benefit applications ranging from human behavior analysis to virtual reality avatars and digital human animation.
Despite advances in \threeD human pose estimation~\cite{baradel2025multi, sun2024aios}, estimated poses often fail to preserve \st, leading to unintended gaps, such as a hand that should touch the face but instead floats in front of it, as shown in \cref{fig:teaser}.
Methods such as \tuch~\cite{muller2021self} and SCP~\cite{fieraru2021learning} have advanced \sct modeling through optimization-based constraints and contact predictions.
However, distinguishing hovering from actual contact in a single RGB image remains challenging, even for humans.
While improvements in 3D human pose estimation have largely been driven by the creation of large-scale training sets, failures in \sct can be attributed to a lack of specialized data, as occlusions and depth ambiguities hinder the generation of \textit{reliable pseudo-ground truth (\pgt)}.
Multi-view capture setups could help, but their complexity and cost make them impractical for large-scale data collection.
To create high-quality training data, we need a scalable method to capture 3D poses involving \sct.\looseness=-1

To that end, we propose a unique multi-modal approach that complements visual pose estimation with \textit{bioimpedance sensing}.
Bioimpedance measures the electrical resistance of a person's body to a small alternating electric current; some scales use this principle to estimate body fat percentage through the feet.
Bioimpedance can be used to detect \st~\cite{sato2012touche,Forte2024-TIM-Bioimpedance}: a person's upper-body bioimpedance can be continuously measured through a dedicated circuit attached to two bracelets worn on the wrists. 
Skin-to-skin contact, such as when the hands touch, creates a parallel electrical pathway through which the current can flow, thus decreasing the total wrist-to-wrist bioimpedance.
We present a signal-processing method that detects these changes in the bioimpedance signal to identify the beginning and end of \sct and reliably distinguish actual contact from close proximity scenarios.
While the sensor detects contact events, we use visual data to infer contact locations, creating a complementary multi-modal approach.

We formalize this integration through \ourmethod (Bioimpedance Timing for Understanding Contact in Humans), a novel framework that refines arm-pose estimates by leveraging bioimpedance measurements as a prior to solve pose ambiguities.
When our sensor indicates that \sct should occur but the inferred 3D pose has no contact, the optimization drives the hand(s) to move toward the body.
We optimize only the arm joints (\ie shoulder, elbow, and wrist) using masked gradient updates.
This targeted approach refines the joints that typically cause hand-initiated contact while leaving the rest of the body unchanged.
Since contact is most difficult to infer along the camera's viewing direction ($z$), our loss function accounts for this high uncertainty by allowing more movement along the $z$-axis.

To evaluate our approach, we introduce a proof-of-concept dataset of synchronized RGB video, bioimpedance measurements, and 3D motion-capture (\mocap) ground truth (\gt) from three subjects.
This in-lab dataset comprises 82 \st gestures derived from observational studies and neuropsychological research.
We also include nine adversarial non-contact gestures to test detection robustness. 
Quantitative analysis demonstrates that our method improves reconstruction accuracy during contact by 11.7\% compared to visual pose estimators alone.
To show applicability outside the lab, we present a miniature wearable bioimpedance sensor and perform an additional validation study; this sensor enables the capture of \sct datasets with everyday clothing in natural indoor and outdoor settings.\looseness=-1

By introducing the scalable and reliable framework of \ourmethod, we advance the field's ability to generate contact-aware training data and more accurately reconstruct natural human movements involving \sct.
Specifically, our contributions include
(1) a novel approach to sensing \sct using bioimpedance;
(2) a novel computer-vision method that integrates this contact information into the 3D human pose estimation task;
(3) a new dataset that includes video with \st gestures, bioimpedance measurements, and \gt 3D poses; and
(4) the design of a miniature bioimpedance sensor that enables practical capture of \sct in the wild.
To encourage further research in this direction, we share our code and dataset for research purposes at \websiteURL.
\section{Related Work}

\qheading{3D Human Pose Estimation with Self-Contact}  
Human pose estimation has advanced significantly with parametric models like SMPL~\cite{Bogo:ECCV:2016} and \smplx~\cite{SMPL-X:2019}.
Recent work has produced robust general-purpose models~\cite{cai2023smpler,sun2024aios,baradel2025multi} and domain-specific solutions~\cite{Forte23-CVPR-SGNify}.
However, all these methods fail to capture proper contact between body parts, primarily due to the fundamental challenge of obtaining reliable training data for poses involving \sct.
SMPLify-XMC~\cite{muller2021self} generates \pgt for the MTP dataset through a ``Mimic-The-Pose'' approach, where volunteers replicate reference poses of 3DCP~\cite{muller2021self} while being photographed;
however, the quality of the resulting dataset depends on how accurately people imitate the depicted poses. Furthermore, 3DCP's contact detection relies on geometric thresholds combining Euclidean and geodesic distances, thus risking confusing close proximity with contact.
Similarly, the \pgt from DSC~\cite{muller2021self} depends on the accuracy of human-annotated contact labels.
These factors influence the performance of the resulting regressor, TUCH~\cite{muller2021self}.
Fieraru \etal~\cite{fieraru2021learning} approach the data problem through manual annotation of two datasets. First, in HumanSC3D, subjects reproduce specific contact types (not full poses) while being recorded with four multi-view RGB cameras, and annotators then manually label the body regions in contact and their correspondence. Second, FlickrSC3D involves manual classification of web images into three categories: contact, no contact, and uncertain contact.
The inclusion of an ``uncertain'' class highlights the inherent difficulty in obtaining definitive \gt for \sct poses from images alone.
Finally, no prior work addresses the temporal dynamics of how \sct states evolve over time.

\qheading{Sensor-Based Detection of Self-Contact}
Recent tactile sensing technologies, such as electronic skin~\cite{xu2023artificial}, enable accurate localization of contact point(s) and precise pressure quantification, providing reliable \pgt for visual data~\cite{mollyn2024egotouch}.
However, the need for extensive sensor coverage across the body limits scalability.
In contrast, compact wearable sensors that measure biomechanical or physiological signals, such as wrist accelerations~\cite{mueller2019self}, electromyography~\cite{mueller2019self}, and hand proximity~\cite{hajika2024radarhand}, offer more practical solutions.
However, their high sensitivity to the movements of the body parts on which they are worn limits their robustness in real-world applications.
Other sensing modalities, such as electrical conductivity, have been used to detect~\cite{zhang2019actitouch} and localize~\cite{zhang2016skintrack} contact but are restricted to interactions occurring near the device.
By contrast, bioimpedance-based approaches offer a scalable solution for detecting skin-to-skin contact; they can operate independently of the contact location~\cite{Forte2024-TIM-Bioimpedance}, requiring only minimal instrumentation~\cite{sato2012touche}.
While this approach has previously been used to classify five static poses~\cite{sato2012touche}, it has not been explored for general detection of skin-to-skin contact or for 3D human pose estimation from video.
Our work leverages the advantages of bioimpedance-based \sct detection to improve the quality of \pgt produced by existing pose estimation methods.
In turn, the refined poses can be used to train more robust pose estimation methods for \sct.

\section{Method}
We introduce \ourmethod, a novel framework that leverages bioimpedance sensing to optimize the 3D arm-pose estimates during \sct events. 

\subsection{Bioimpedance Sensing}

\begin{figure}[tb]%
    \centering
    \includegraphics[width=0.6\columnwidth]{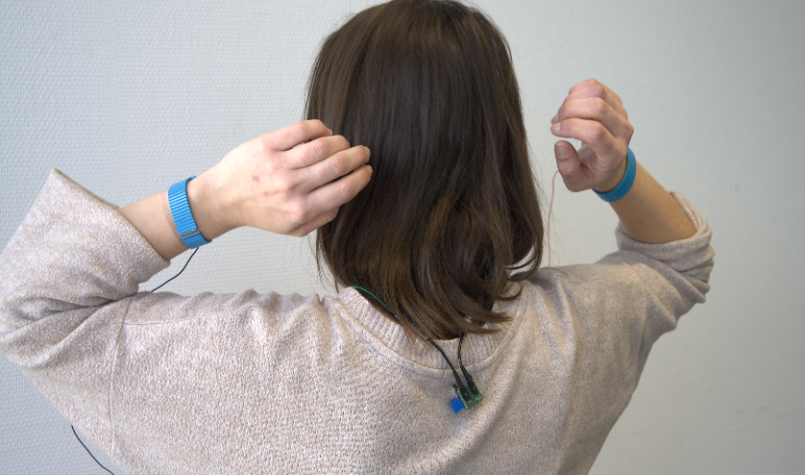}
    \vspace{-0.1in}
    \captionof{figure}{Wearable setup. We created a miniaturized sensor that consists of a processor, shown here attached to the wearer's shirt, and two electrodes embedded in the blue bracelets on their wrists. All components can be hidden beneath clothing.}%
    \label{fig:wearable}
\end{figure}

While bioimpedance sensing can detect \sct between any body parts with appropriate electrode placement, we focus on \st due to its prevalence.
Following~\cite{Forte2024-TIM-Bioimpedance}, we use a high-quality commercial impedance analyzer (MFIA, Zurich Instruments) with a wrist-to-wrist electrode configuration.
However, this device's dimensions (29\,cm $\times$ 23\,cm $\times$ 10\,cm), weight ($\sim$4\,kg), and cost ($\sim$15,000 USD) make it impractical for large-scale data collection. Thus, we create a compact, low-cost, and wearable version~\cite{Forte2025-Patent} (see~\cref{fig:wearable}).
Our custom device is small (2\,cm $\times$ 1.8\,cm $\times$ 1.1\,cm), lightweight (0.02\,kg), affordable ($\sim$20 USD), and easy to build and use.
It operates on a small battery that allows over three hours of continuous use.
The core component is a microcontroller (Adafruit QT Py ESP32-C3) mounted on a custom circuit board and connected via thin wires to two electrodes. To achieve secure and comfortable mounting on the wrists, we employ commercial electrode wristbands commonly used to prevent electrostatic discharge.
This miniaturized device makes it possible to detect \sct even when the subject is outdoors and wearing everyday clothing, demonstrating the feasibility of a more wearable and scalable solution.\looseness=-1

\subsection{\ourmethod}
\ourmethod consists of two steps: (1) detecting self-contact using the bioimpedance signal and (2) optimizing the arm pose to enforce contact, if detected.

\subsubsection{Self-Contact Detection}
\begin{figure}[tb]%
    \centerline{   \includegraphics[width=\columnwidth,clip=true,trim=0cm 0.3cm 0 0]{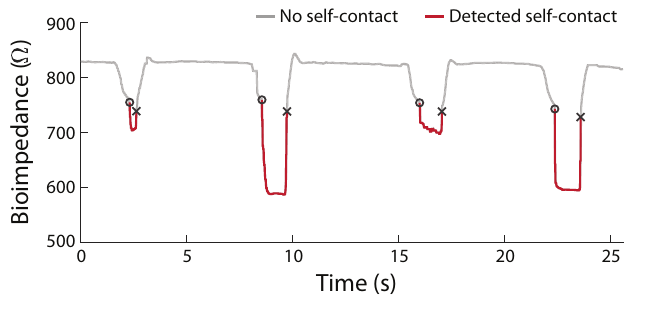}}
    \vspace{-0.1in}
    \captionof{figure}{Sample bioimpedance signal showing four self-contact gestures over time. The bioimpedance magnitude begins to change when the person moves from the resting pose. The start of each self-contact triggers an abrupt drop in bioimpedance that is detected by our algorithm and marked with $\circ$. The thick red segments show the duration of the self-contacts, and their estimated ends are marked with $\times$.}%
    \label{fig:biosignal}
\end{figure}

As shown in \cref{fig:biosignal}, wrist-to-wrist bioimpedance decreases sharply during instances of skin-to-skin contact. 
For \sct detection, the signal is first resampled via linear interpolation to ensure a fixed sampling rate, then smoothed using a median filter with a window size of 100\,ms, and finally differentiated. 
The beginning of each \sct event is identified using an adaptive thresholding method applied to the processed signal, similar to techniques used in audio onset detection~\cite{bock2013maximum} and in detecting movement or muscle contraction~\cite{carvalho2023review, hodges1996comparison}.
Specifically, we set the self-contact threshold to approximately one-third of the average of the three lowest minima in the processed recording; this thresholding procedure was designed and refined empirically.
The end of each \sct is defined as the time at which the bioimpedance magnitude returns to 98\% of its pre-contact value.
To reduce false positives, we apply temporal constraints informed by prior work on \sct dynamics~\cite{mueller2019self}. 
Our approach prioritizes specificity over sensitivity, allowing more false negatives (\ie missed contacts) than false positives (\ie incorrect contact detections) in order to reduce the risk of incorrectly optimizing non-contact poses.
Finally, the detected contact events are converted into a binary touch sequence aligned with the video frames.

\subsubsection{Arm-Pose Optimization}
We adopt the \smplx model~\cite{SMPL-X:2019} to represent the 3D body and output a \threeD body mesh. 
\smplx is a differentiable function, $M(\theta, \beta, \psi)$, parameterized by body pose $\theta$, including finger articulations (represented by the hand pose $\theta_h$), body shape $\beta$, and facial expressions $\psi$.
We first run an off-the-shelf method (\eg \multihmr~\cite{baradel2025multi}) to obtain the initial \smplx estimates for a video sequence.
We then average the estimated $\beta$ and do not optimize body shape to maintain a consistent body shape across the sequence.
In the first frame, we optimize body translation and global orientation (\ie pelvis joint) following \smplifyx~\cite{SMPL-X:2019} and use the estimated values for the full sequence.
We do not optimize the body pose for frames in which the bioimpedance-based sensor detects no contact.
For frames with detected contact, \ourmethod refines the arm poses through the following contact-aware optimization.
For each hand, it identifies potential contact regions by computing distances between hand vertices $\mathbf{v}$ and target vertices $\mathbf{u}$ on the input \smplx mesh. Target vertices include all upper-body vertices (including the head, face, and opposite hand) apart from the hand's own arm.
Since existing methods perform reasonably well in the camera's image plane ($x$- and $y$-coordinates) but struggle with depth ($z$-axis), distances along the camera's $z$-axis are weighted less, \ie a 1\,cm error along $z$ is equivalent to a 0.25\,cm error in the viewing plane.
This weighting is based on quantitative analysis that showed that errors along the camera's viewing direction are typically 3--4 times larger.
From these weighted distances, we determine the closest vertex pair(s):
$\mathcal{P}_h = \{(\mathbf{v}_i, \mathbf{u}_i)\}$.
When the hand is close to multiple body regions (\eg it could touch both the face and the opposite hand), $\mathcal{P}_h$ contains multiple vertex pairs from different regions that are combined during optimization to avoid discontinuities.
The initial 3D positions of the vertices of $\mathcal{P}_h$ are stored and used to maintain spatial consistency during optimization.

The weighted vertex-pair distances of both hands are then compared to determine which arms to optimize: both when these distances differ by at most 50\% of the smaller distance, and otherwise only the arm with the smaller distance.
Given the selected arm(s), the relevant joint parameters (\ie shoulder, elbow, and wrist) are optimized through masked gradient updates:
\begin{equation}
    \boldsymbol{\theta}_{i+1} = \boldsymbol{\theta}_i - \eta\nabla_{\boldsymbol{\theta}} \mathcal{L} \odot \mathbf{M}_a
\end{equation}
where $\boldsymbol{\theta}_i \in \mathbb{R}^{63}$ represents the \smplx body pose parameters in axis-angle representation at iteration $i$, $\eta$ is the learning rate, $\nabla_{\boldsymbol{\theta}} \mathcal{L}$ is the gradient of the loss function with respect to all pose parameters, and $\mathbf{M}_a \in \{0,1\}^{63}$ is the binary mask.  
The loss function through which we update the arm pose is:
\begin{equation}
    \mathcal{L} = \mathcal{L}_{\text{2D}} + \lambda_{\text{contact}} \mathcal{L}_{\text{contact}}
\end{equation}
where $\mathcal{L}_{\text{2D}}$ ensures consistency with the 2D joint observations of the arms, and
\begin{equation}
\mathcal{L}_{\text{contact}} = \mathcal{L}_{\text{consistency}} + \mathcal{L}_{\text{interpenetration}} + \mathcal{L}_{\text{proximity}} 
\end{equation}
$\mathcal{L}_{\text{consistency}}$ preserves the initial spatial relationships, $\mathcal{L}_{\text{interpenetration}}$ prevents mesh interpenetration through a barrier function, and $\mathcal{L}_{\text{proximity}}$ encourages contact convergence.
Specifically,
\begin{equation}
\mathcal{L}_{\text{proximity}} = \sum_{h \in \mathcal{H}} \sum_{(\mathbf{v}_i,\mathbf{u}_i) \in \mathcal{P}_h} \sum_{d \in \{x, y, z\}} \omega_d |v_i^d - u_i^d|
\end{equation}
where $\mathcal{H}$ is the set of active hands and $\omega_x$, $\omega_y$, and $\omega_z$ are camera-adaptive weights (with depth weighted four times higher than the viewing plane).
Optimization continues until contact is achieved (distances $\leq$ 5\,mm in all axes) or iteration limits are reached.
This approach achieves physically plausible \sct while maintaining consistency with both 2D keypoint observations and the initial pose estimates.
Finally, for temporal consistency, we apply OneEuroFilter-based smoothing in post-processing, as in~\cite{VIBE:CVPR:2020} 
(see supplementary video).

\section{Dataset}
\begin{figure*}[tb!]%
    \centerline{    \includegraphics[width=\textwidth,clip=true,trim=0.02cm 0cm 0.02cm 0cm]{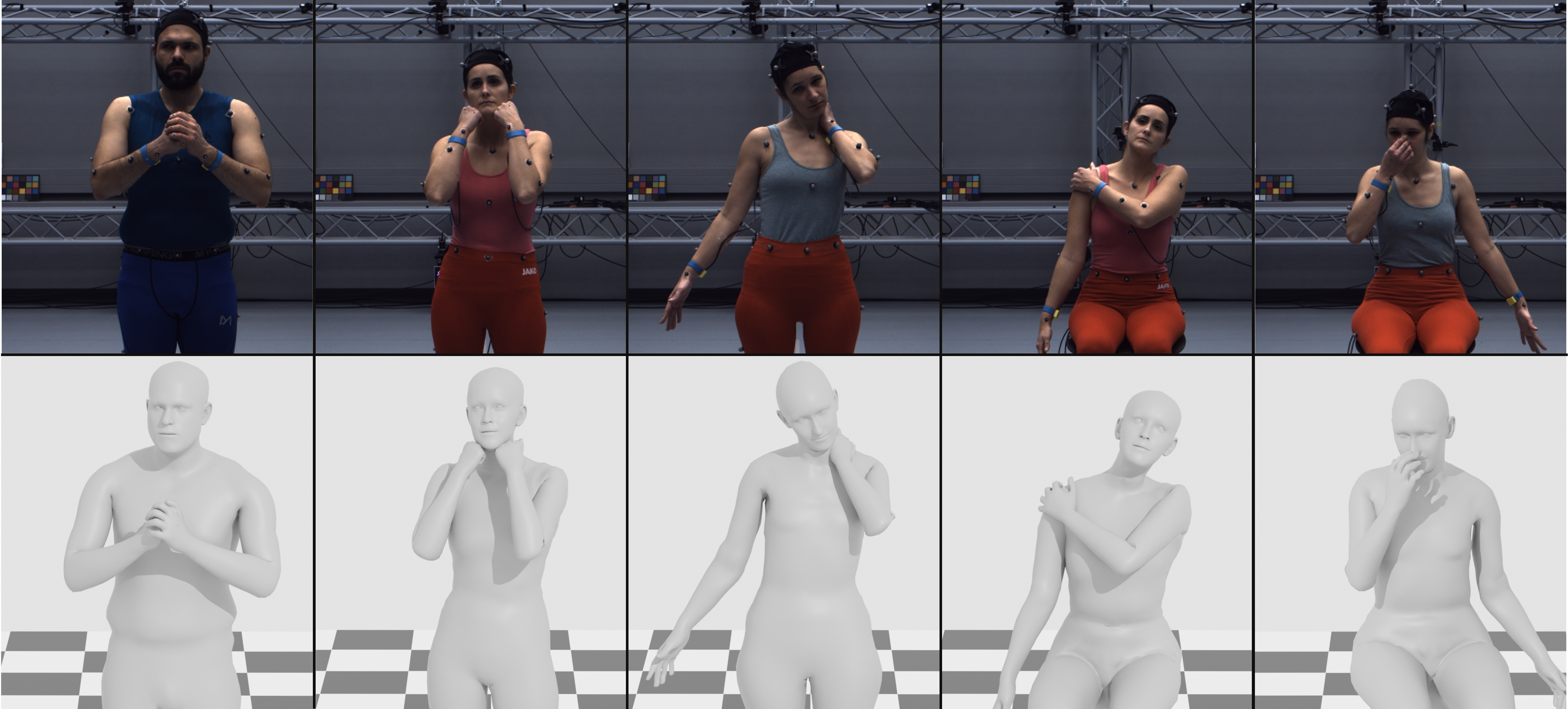}}
    \vspace{-0.1in}
    \captionof{figure}{Sample contacts that cause a significant change in the bioimpedance of the user. Any self-contact (direct on skin or through body hair) greatly impacts the wrist-to-wrist bioimpedance signal. The videos from which these images were taken and their ground-truth \smplx meshes are part of our newly collected dataset.}%
    \label{fig:example_contact}
\end{figure*}
To evaluate \ourmethod, we collected a new dataset of synchronized frontal RGB videos, bioimpedance measurements, and 3D motion capture. \Cref{fig:example_contact} shows five sample poses from this dataset, representing the first collection of directly sensed contact-aware training data suitable for improving pose estimation models.
The experimental procedure was reviewed and approved by our local ethics board. 

The dataset comprises 82 \textit{common dynamic self-touch gestures} derived from observational studies, neuropsychological research, and poses from the MTP dataset~\cite{muller2021self}. These gestures are organized into the following categories according to the body parts in contact:
(a) 38 hand-to-face or hand-to-head poses (\eg temple touch, nose touch, hands covering the face) reflecting attentional refocusing~\cite{mueller2019self} or emotional regulation~\cite{harrigan1985self, pang2022individual};
(b) 12 hand-to-hand gestures (\eg hand wringing, hands clasped, fingers interlocked) associated with emotional regulation~\cite{muller2021self, pang2022individual} and cognitive load~\cite{muller2021self}; 
(c) 11 hand-to-neck or hand-to-shoulder movements (\eg shoulder rub) linked to self-support and stress relief~\cite{harrigan1985self};
(d) nine hand-to-chest or hand-to-torso movements (\eg palm over heart, self-hugs) often associated with self-reassurance or stress relief~\cite{dreisoerner2021self}; 
(e) eight behind-the-back gestures (\eg holding hands or clasping the opposite forearm behind the back) related to postural control, tension release, or self-soothing (note that these contacts are not visible in the frontal RGB camera); 
(f) four hands-to-body gestures, where both hands contact different body parts simultaneously (\eg one hand supporting the head and the other touching the other arm), typical of postural support.
We also include nine adversarial non-contact gestures that can be visually interpreted as \st contacts (\eg hand close to the forehead or mouth) when seen in a frontal view.
For unilateral gestures, we include variants performed with each hand.

\begin{table}[t!]
    \centering
    \renewcommand{\tabularxcolumn}[1]{>{\RaggedRight}p{#1}} 
    \resizebox{\linewidth}{!}{
        \renewcommand{\arraystretch}{1.6} 
\large
\begin{tabular}{@{}l c c c c@{}}
\toprule
\textbf{Dataset} & \textbf{\makecell{\# of Contact\\Frames}} & \textbf{Data Type} & \textbf{Motion} & \textbf{Annotation}\\
\midrule
3DCP~\cite{muller2021self} & 1,653 &\makecell{\smplx\\(from 3D scans, 3D \mocap)}& Static & \makecell{Automatic\\(geometry)}\\
\midrule
MTP~\cite{muller2021self} & 3,731 & \makecell{RGB images,\\(fitted) \smplx} & Static & \makecell{Automatic\\(geometry)}\\
\midrule
DSC~\cite{muller2021self} &30,000& RGB images & Static & Manual\\
\midrule
HumanSC3D~\cite{fieraru2021learning}& 4,128 &\makecell{3D \mocap,\\multi-view RGB images}&Dynamic& Manual\\
\midrule
FlickrSC3D~\cite{fieraru2021learning}& 3,969 & RGB images & Static & Manual\\
\midrule
Ours &19,183&\makecell{\smplx (from 3D \mocap),\\RGB images, bioimpedance}&Dynamic&\makecell{Automatic\\(bioimpedance)}\\
\bottomrule
\end{tabular}
    }
    \vspace{-0.1in}
    \caption{Comparison of datasets for \sct.}
    \label{tab:dataset}
\end{table}

Three participants (two females and one male) took part in the study.
Although the number of participants is limited, our electrode configuration and bioimpedance sensing parameters are based on Forte \etal~\cite{Forte2024-TIM-Bioimpedance}, who validated this sensing method across a diverse population that varied in body mass index, sex, ethnicity, and handedness.
Based on their results, we measured the bioimpedance magnitude using an alternating current signal at 2.29\,MHz, with data sampled at 13.4\,kHz.
The participants were captured with a Vicon mocap system at 30 fps while wearing the two electrode bracelets.
The mocap system was synchronized with the impedance analyzer (MFIA, Zurich Instruments) and a frontal 1300$\times$1400 RGB camera at 30 fps, framing a full-body view. 
All participants performed the gestures while standing; two participants repeated them while sitting.
The arms rested at the participant's sides at the beginning and end of each gesture. 
\gt \smplx meshes were obtained by scanning the participants in a 4D body scanner in several poses and fitting the \smplx model to these scans.
A personalized body-shape mesh of each participant was obtained by averaging the \smplx meshes in the canonical pose. 
MoSh++ was then used to fit this mesh to the mocap markers~\cite{mahmood2019amass}, providing an accurate \gt 3D body mesh for each frame (\cref{fig:example_contact}).
\Cref{tab:dataset} compares our dataset with existing datasets related to self-contact.
\section{Experiments}

\subsection{Evaluation of Self-Contact Detection}
To quantitatively evaluate the contact-detection accuracy of our sensing approach, we manually labeled binary \sct over time for 57 gestures (including adversarial non-contact gestures); these were randomly selected from all three participants and both sitting and standing poses, excluding gestures used for setting the algorithm's parameters.
It is important to note that both frames and bioimpedance were used for manual labeling, as frames alone were not reliable in several cases.
The algorithm has a sensitivity of 0.858 and specificity of 0.992 (false-negative rate of 0.142 and false-positive rate of 0.008).
These high sensitivity and specificity values confirm the reliability of the sensor as a proxy for \gt in contact detection.
Furthermore, none of the non-contact gestures were misclassified as contact.

\subsection{Quantitative Evaluation}

\begin{table*}[t!]
\centering

\begin{tabular}{l >{\centering\arraybackslash}p{2cm} >{\centering\arraybackslash}p{2cm} >{\centering\arraybackslash}p{2cm} >{\centering\arraybackslash}p{2cm} >{\centering\arraybackslash}p{2cm} >{\centering\arraybackslash}p{2cm}}
\toprule
\textbf{Method} & \makecell{PA-V2V $\downarrow$\\(mm)} & \multicolumn{3}{c}{\makecell{Euclidean Joint Location Error $\downarrow$\\(mm)}} & \multicolumn{1}{c}{\makecell{Detection Rate $\uparrow$\\(\%)}} & \multicolumn{1}{c}{\makecell{V-Distance $\downarrow$\\(mm)}}\\
\cmidrule(lr){3-5}
& & Shoulder & Elbow & Wrist & & \\
\midrule
\multihmr \cite{baradel2025multi} & 57.46 & \textbf{23.49} & \textbf{29.86} & 65.37 & 41.28 & 87.48\\
+ 2D Loss & 77.41 & 26.74 & 39.31 & 84.97 & 42.16 & 124.66\\
+ Contact Loss & 50.74 & 23.71 & 31.71 & 57.09 & \textbf{79.35} & 72.59\\
+ \ourmethod & \textbf{50.21} & 23.95 & 30.99 & \textbf{56.29} & 78.34 & \textbf{71.22} \\
\midrule
\aios \cite{sun2024aios} & 72.24 & \textbf{21.84} & \textbf{39.83} & 77.29 & 45.87 & 99.02\\
+ 2D Loss & 86.61 & 27.01 & 41.31 & 93.38 & 41.58 & 130.64\\
+ Contact Loss & 65.10 & 22.13 & 43.99 & 68.29 & \textbf{80.60} & 82.01\\
+ \ourmethod & \textbf{62.79} & 22.42 & 40.65 & \textbf{65.61} & 78.48 & \textbf{79.16}\\
\midrule
\tuch \cite{muller2021self} & 70.55 & \textbf{28.24} & 41.03 & 58.71 & 59.46 & 96.31\\
+ 2D Loss & 77.12 & 33.66 & 49.00 & 77.03 & 49.52 & 110.46 \\
+ Contact Loss & 64.95 & 28.24 & 40.65 & 53.24 & \textbf{85.25} & 86.95 \\
+ \ourmethod & \textbf{63.99} & 28.27 & \textbf{40.59} & \textbf{52.91} & 84.60 & \textbf{86.26}\\
\bottomrule
\end{tabular}

\caption{Performance evaluation reporting vertex-to-vertex arm mesh error (PA-V2V), Euclidean error for each arm joint, and the contact metrics of detection rate and distance between the contact vertices in \gt and in the estimate. We compare each off-the-shelf method with its extension with \ourmethod, including ablation variants isolating the effects of \ourmethod's two losses: 2D loss and contact loss. We bold the best result of these four methods for each metric. $\downarrow$ lower value means better performance. $\uparrow$ higher value means better performance.
}
\vspace{-0.1in}
\label{tab:v2v}
\end{table*}

We quantitatively evaluate \ourmethod, comparing vanilla \multihmr~\cite{baradel2025multi}, \aios~\cite{sun2024aios}, and \tuch~\cite{muller2021self} (converted to \smplx) with their \ourmethod extensions.
\multihmr and \aios are the two state-of-the-art methods in human pose estimation, and \tuch is tuned for \sct.
The evaluation uses synchronized meshes from Vicon markers~\cite{mahmood2019amass} as \gt and is conducted on a per-frame basis before any temporal smoothing. 
All methods estimate \smplx meshes with the same topology, enabling Procrustes alignment (PA) for fair comparison.  
We compute the mean per-vertex error (\PAvtov) of the optimized full arm regions and the Euclidean errors of the key arm joints, as well as the contact accuracy.
Specifically, we compare each method's output with the \gt meshes to compute detection rate (\ie how often a \gt contact is reconstructed) and contact preservation (\ie how well spatial relationships between contacting vertices are maintained) by measuring the Euclidean distances between the same vertex pairs that were in contact in the \gt.
\Cref{tab:v2v} reports the results of these metrics for the three input methods, including an ablation study showing the individual contributions of our loss components (2D loss and contact loss). 

Among the input methods, \aios has the highest \PAvtov error (72.24\,mm), while \multihmr achieves the best baseline performance (57.46\,mm).
As expected, \tuch has the highest contact detection rate (59.46\%) among the baselines, reflecting its specialization for \sct scenarios.
In general, \ourmethod consistently improves all input methods across key metrics. 
For \PAvtov error, we achieve an average 11.7\% improvement.
Joint-level analysis reveals varied improvements across the arm: shoulder and elbow joint errors show overall a slight degradation (0.36\,mm and 0.50\,mm), while wrist joint locations drastically improve (8.85\,mm). 
\ourmethod increases contact detection rates considerably across all methods, with an average improvement of 31.60 percentage points and an average reduction of 15.39\,mm in distances between contacting vertex pairs. These findings indicate that our method not only reconstructs contacts more reliably but also maintains better spatial relationships between contacting surfaces.

The ablation study reveals that while combining 2D and contact losses improves baseline performance, the contact loss provides the dominant contribution.
The 2D loss (\smplifyx's re-projection error) shows minimal benefit compared to our contact-specific loss component.
The highest detection rate is achieved when adding the contact loss alone, as it is not constrained by keypoint positions; however, this ablation preserves contact less well, as the optimization prioritizes achieving contact and does not preserve alignment with the original keypoints.
This result confirms that bioimpedance provides unique information that cannot be derived from 2D keypoints alone, validating our multi-modal approach of prioritizing contact-aware optimization.
Since \multihmr demonstrates the most reliable baseline performance, it achieves the best overall results when enhanced with \ourmethod, reaching 50.21\,mm \PAvtov error.

\begin{figure}[b!]%
    \vspace{-0.1in}
    \centering
    \includegraphics[width=0.6\linewidth]{figure/comparison_combined_zoom_250.0mm_clean.pdf}
    \vspace{-0.1in}
    \caption{Robustness evaluation. The plot shows PA-V2V errors before ($x$-axis) and after ($y$-axis) applying \ourmethod to the three input methods. Points below the dashed black line indicate improvement. The $R^2$ values show how well the fitted curves explain the variance in the data.}
    \label{fig:initialization}
\end{figure}
Finally, we conduct a robustness evaluation to investigate \ourmethod's effectiveness across different levels of input reconstruction error. \Cref{fig:initialization} shows the relationship between the PA-V2V errors of the input methods and the corresponding errors after applying \ourmethod.
The analysis reveals that \ourmethod consistently improves all baseline methods across the entire error spectrum, with the most pronounced improvements in the 50--150\,mm range.

\subsection{Qualitative Evaluation}

\begin{figure}[!t]%
    \centering
    \vspace{-0.1in}
    \includegraphics[width=\columnwidth]{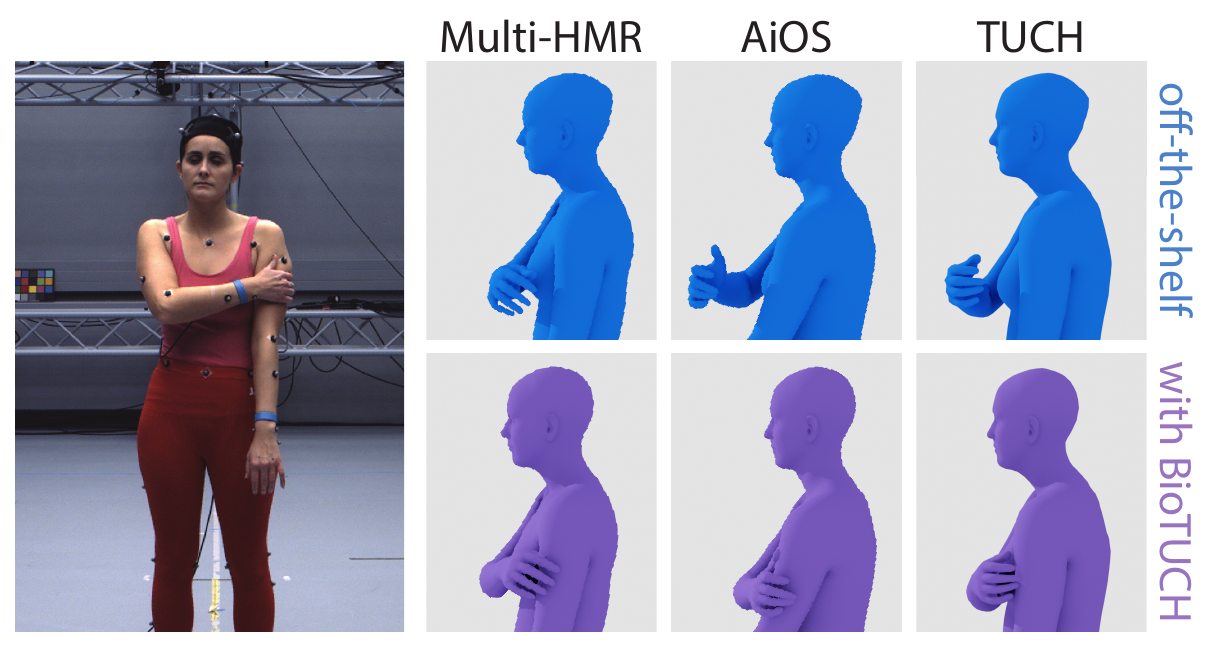}
    \vspace{-0.2in}
    \caption{Qualitative results. The left column shows an RGB image from a sample self-contact gesture in our dataset. On the right, the estimates of the off-the-shelf methods (\ie \multihmr, \aios, \tuch) are shown in blue in the first row, and their \ourmethod extensions are shown in purple in the second row. \ourmethod clearly optimizes the arm joints to bring the hand into self-contact at the observed location in the image plane.}
    \label{fig:qualitative}
\end{figure}

We qualitatively evaluated the three input methods and their \ourmethod extensions on our dataset.
Among the input methods, we noticed that \tuch performs particularly well for hand-to-face contacts.
\multihmr and \aios instead do not exhibit any significant variations across different body parts.
Overall, as expected, the most accurate contact predictions for all input methods occur when the contact is along the camera's $xy$ plane (\eg a hand touching the temple), and the most challenging scenarios are the behind-the-back gestures.
In both cases, \ourmethod has minimal impact.
In case of contacts in the $xy$ plane, the contact is often already reconstructed by the input methods, so \ourmethod does not optimize the 3D pose. Conversely, the significant errors of the estimated joint positions of the behind-the-back gesture do not allow meaningful contacts.
\ourmethod proves particularly beneficial for visible contacts that occur in line with the camera's optical axis. Although these contacts are visible, they pose challenges for the off-the-shelf methods due to the depth ambiguity problem. \Cref{fig:qualitative} illustrates the estimates for a sample gesture from \multihmr, \aios, and \tuch, along with the improvements achieved by \ourmethod. 
Overall, \ourmethod effectively optimizes arm poses to ensure contact while maintaining the consistency of 2D points. However, the optimization halts once contact is achieved. Consequently, inaccuracies in the finger articulation of the input method can affect the stopping condition of \ourmethod, as seen for \tuch in \Cref{fig:qualitative}.
More qualitative results are shown in the supplemental video.

\begin{figure}[tbp]%
    \centering
    \includegraphics[width=0.99\columnwidth]{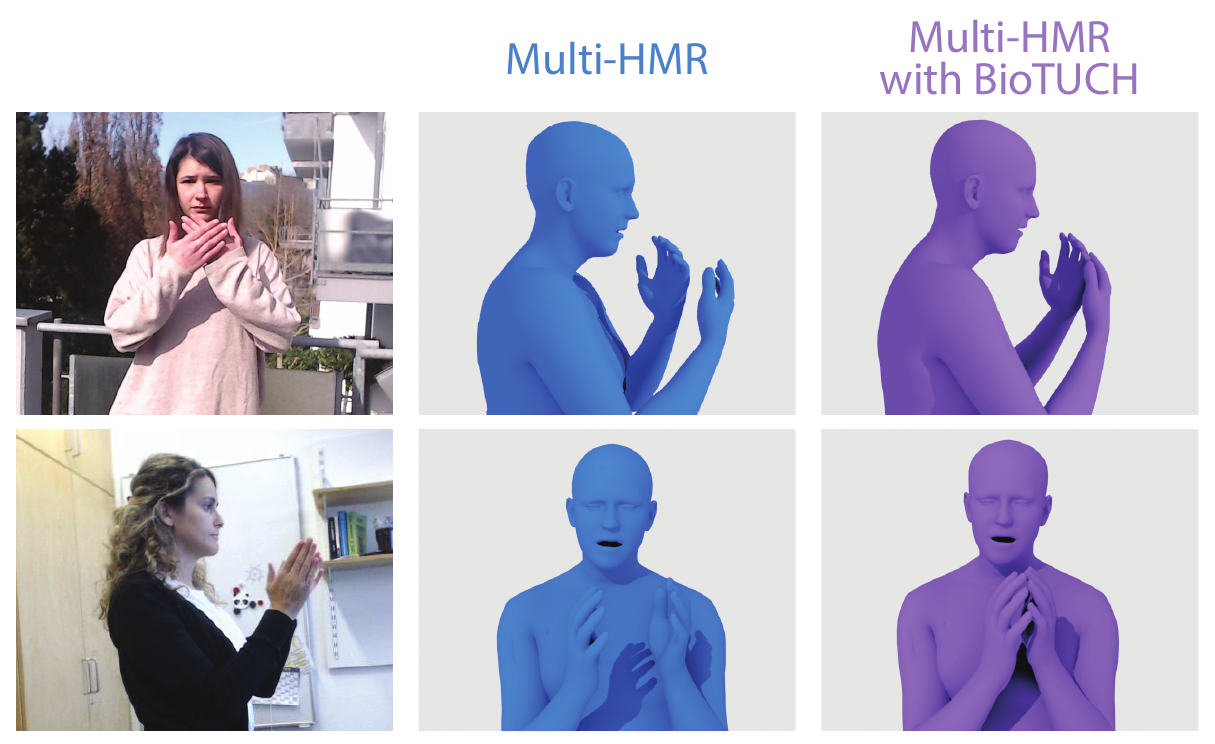}
    
    \vspace{-0.1in}
    \caption{Results of two in-the-wild captures. When \multihmr fails to reconstruct a self-contact detected by our miniature bioimpedance sensor, \ourmethod optimizes the arm joints to resolve depth ambiguity and enforce contact.}
    \label{fig:inthewild}
\end{figure}

\subsection{In-the-wild Capture}
To demonstrate outside-the-lab feasibility, we collected two short in-the-wild sequences using our miniature sensor (see \cref{fig:wearable}) worn beneath clothing and a single frontal or side-view camera.
The bioimpedance data and the video were recorded and automatically synchronized using MATLAB.
As shown in \cref{fig:inthewild}, \multihmr fails to accurately reconstruct \sct along the camera axis in both cases. Whether the camera is viewing the front or side of the user, \ourmethod addresses the depth ambiguity by optimizing the arm pose, successfully generating the detected contact.
\section{Discussion}

\begin{figure}[!b]
    \centering
    \vspace{-0.15in}
    
    \includegraphics[width=0.99\linewidth]{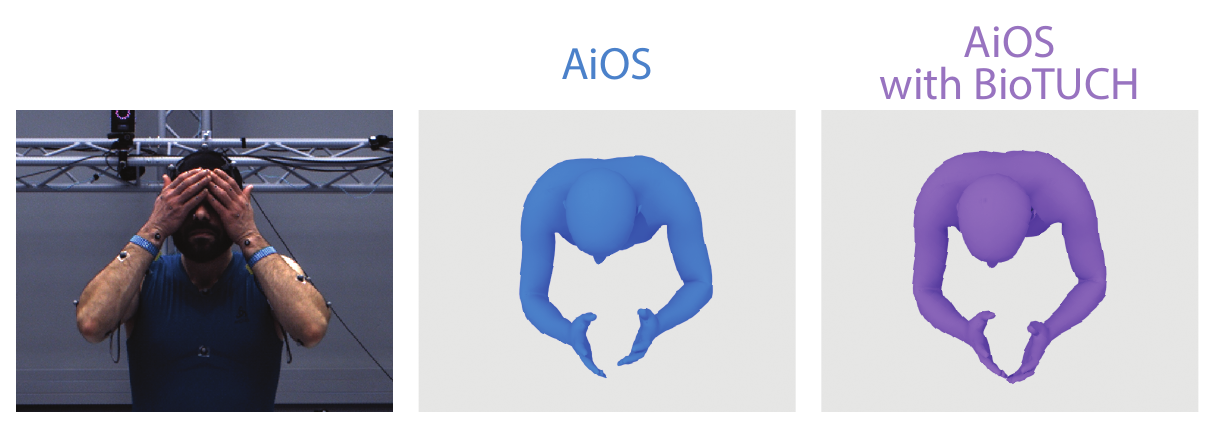}
    \vspace{-0.1in}
    
    \caption{Input RGB image for a challenging two-handed \st gesture, with top-down views of the estimates of \aios and \aios with \ourmethod.}
    \label{fig:failure1}
\end{figure}
While \ourmethod demonstrates significant improvements in contact-aware pose estimation, several areas present opportunities for future enhancement.
Our current approach relies on the input mesh estimates to determine which parts of the body should be in contact.
However, as shown in \cref{fig:failure1}, input meshes can exhibit substantial errors along all axes, leading to suboptimal contact region identification.
Additionally, when a single hand contacts multiple body parts simultaneously (as in~\cref{fig:failure1}, where each hand touches both the other hand and the face), distance-based region identification is fundamentally limited by the initial mesh's spatial configuration. For instance, in \cref{fig:failure1}, errors along the $z$-axis are significantly larger than those in the $xy$-plane, and \ourmethod prioritizes optimizing the hand-to-hand contact over the hand-to-face contacts.
Even localized inaccuracies, such as incorrect finger articulation, can stop the optimization too early, as observed in the bottom row of \cref{fig:inthewild}.
Future work should address these limitations by using visual evidence to identify the contact regions when contact is visible in the raw input images, thus allowing the optimization to maintain all visually identified contacts rather than selecting only the closest mesh-based estimate.

Beyond these improvements, \ourmethod can be readily extended in several directions. We focused on hand-initiated contacts as they are most common, but bioimpedance sensing can be used to detect \sct at any location by repositioning the electrodes (\eg moving them to the ankles for lower-limb contact detection) and updating the active body parts and target vertices.
Furthermore, we used binary contact detection; however, bioimpedance signals can provide richer information, including the location and size of the contact~\cite{Forte2024-TIM-Bioimpedance}. This sensing approach can also detect skin-to-clothing contact through fabric when both fabric sides contact the skin~\cite{Forte2024-TIM-Bioimpedance}, as we observed in some hand-to-chest contacts in our dataset.
Finally, large-scale collection with our miniature sensor could provide improved training data for methods that regress body parameters from RGB images.
The next step would be to collect such a dataset and train a regressor on bioimpedance-corrected meshes, similar to how \tuch~\cite{muller2021self} trained on visually derived \pgt.

\section{Conclusions}
We present \ourmethod, a novel framework to improve reconstruction of self-contact in \threeD avatars from monocular \rgb video.
Quantitative and qualitative results show that
\ourmethod improves contact reconstruction in all tested off-the-shelf methods by leveraging direct-contact sensing.
Our wearable bioimpedance sensor, signal processing, and pose optimization represent a step toward capturing accurate \pgt with self-contact at scale. 
We showed that contact can be reliably detected using bioimpedance, we built and tested a wearable sensor, we quantitatively validated our method in an instrumented lab setting, and we showed feasibility in the wild.
The presented framework enables the capture of new datasets that can increase the accuracy of \threeD human pose estimation involving self-contact.

{
\vspace{-0.7em}
\paragraph*{Acknowledgments}
{We thank Tsvetelina Alexiadis for trial coordination; Markus Höschle for the capture setup; Taylor McConnell for data-cleaning coordination; Florentin Doll, Arina Kuznetcova, Tomasz Niewiadomski, and Tithi Rakshit for data cleaning; Giorgio Becherini for MoSh++; and IMPRS-IS for supporting Maria-Paola Forte and Jan Ulrich Bartels. 
}}

\vspace{1ex}{
    \qheading{Disclosure}
    While Michael J. Black is a co-founder and Chief Scientist at Meshcapade, his research in this project was performed solely at, and funded solely by, the Max Planck Society.
}

%
%
{
\bibliographystyle{config_template/ieee_fullname}
\bibliography{config_custom/BIB}

@String(CVPR 	=	{{Computer Vision and Pattern Recognition (CVPR)}})

@String(ECCV 	=	{{European Conference on Computer Vision (ECCV)}})

@STRING(AAAI	=	{{AAAI Conference on Artificial Intelligence}})

@inproceedings{muller2021self,
  title     = {On Self-Contact and Human Pose},
  author    = {M{\"u}ller, Lea and Osman, Ahmed A. A. and Tang, Siyu and Huang, Chun-Hao P. and Black, Michael J.},
  booktitle = CVPR,
  pages     = {9990--9999},
  year      = {2021}
}

@inproceedings{SMPL-X:2019,
  title     = {Expressive Body Capture: {3D} Hands, Face, and Body from a Single Image},
  author    = {Pavlakos, Georgios and Choutas, Vasileios and Ghorbani, Nima and Bolkart, Timo and Osman, Ahmed A. A. and Tzionas, Dimitrios and Black, Michael J.},
  booktitle = CVPR,
  pages     = {10975--10985},
  year      = {2019}
}

@inproceedings{Bogo:ECCV:2016,
  title         = {Keep it {SMPL}: {A}utomatic Estimation of {3D} Human Pose and Shape From a Single Image},
  author        = {Bogo, Federica and Kanazawa, Angjoo and Lassner, Christoph and Gehler, Peter and Romero, Javier and Black, Michael J.},
  booktitle     = ECCV,
  pages         = {561--578},
  series        = {Lecture Notes in Computer Science},
  publisher     = {Springer International Publishing},
  month         = oct,
  year          = {2016},
  doi           = {},
  month_numeric = {10}
}

@article{mueller2019self,
  title={Self-touch: {C}ontact durations and point of touch of spontaneous facial self-touches differ depending on cognitive and emotional load},
  author={Mueller, Stephanie Margarete and Martin, Sven and Grunwald, Martin},
  journal={PLOS ONE},
  volume={14},
  number={3},
  pages={e0213677},
  year={2019},
  publisher={Public Library of Science San Francisco, CA USA}
}

@article{harrigan1985self,
  title={Self-touching as an indicator of underlying affect and language processes},
  author={Harrigan, Jinni A},
  journal={Social Science \& Medicine},
  volume={20},
  number={11},
  pages={1161--1168},
  year={1985},
  publisher={Elsevier}
}

@inproceedings{sato2012touche,
  title={Touch{\'e}: {E}nhancing touch interaction on humans, screens, liquids, and everyday objects},
  author={Sato, Munehiko and Poupyrev, Ivan and Harrison, Chris},
  booktitle={Proceedings of the ACM Conference on Human Factors in Computing Systems (CHI)},
  pages={483--492},
  year={2012},
  publisher={ACM},
  address={Austin, Texas, USA}
}

@inproceedings{Forte2024-TIM-Bioimpedance,
  title={Wrist-to-Wrist Bioimpedance Can Reliably Detect Discrete Self-Touch},
  author={Forte, Maria-Paola and Vardar, Yasemin and Javot, Bernard and Kuchenbecker, Katherine J.},
  booktitle={{IEEE Transactions on Instrumentation and Measurement}},
  year={2025}
}

@inproceedings{Forte23-CVPR-SGNify,
  title = {Reconstructing Signing Avatars from Video Using Linguistic Priors},
  author = {Forte, Maria-Paola and Kulits, Peter and Huang, Chun-Hao Paul and Choutas, Vasileios and Tzionas, Dimitrios and Kuchenbecker, Katherine J. and Black, Michael J.},
  booktitle = {IEEE/CVF International Conference on Computer Vision and Pattern Recognition (CVPR)},
  pages = {12791--12801},
  month = jun,
  year = {2023},
  doi = {},
  month_numeric = {6}
}

@inproceedings{baradel2025multi,
  title={Multi-{HMR}: Multi-person whole-body human mesh recovery in a single shot},
  author={Baradel, Fabien and Armando, Matthieu and Galaaoui, Salma and Br{\'e}gier, Romain and Weinzaepfel, Philippe and Rogez, Gr{\'e}gory and Lucas, Thomas},
  booktitle={European Conference on Computer Vision},
  pages={202--218},
  year={2025},
  organization={Springer}
}

@inproceedings{sun2024aios,
  title={{AiOS}: All-in-One-Stage Expressive Human Pose and Shape Estimation},
  author={Sun, Qingping and Wang, Yanjun and Zeng, Ailing and Yin, Wanqi and Wei, Chen and Wang, Wenjia and Mei, Haiyi and Leung, Chi-Sing and Liu, Ziwei and Yang, Lei and others},
  booktitle={IEEE/CVF Conference on Computer Vision and Pattern Recognition (CVPR)},
  pages={1834--1843},
  year={2024}
}

@inproceedings{fieraru2021learning,
  title={Learning complex {3D} human self-contact},
  author={Fieraru, Mihai and Zanfir, Mihai and Oneata, Elisabeta and Popa, Alin-Ionut and Olaru, Vlad and Sminchisescu, Cristian},
  booktitle={Proceedings of the AAAI Conference on Artificial Intelligence (AAAI)},
  pages={1343--1351},
  year={2021}
}

@article{xu2023artificial,
  title={Artificial intelligence-powered electronic skin},
  author={Xu, Changhao and Solomon, Samuel A and Gao, Wei},
  journal={Nature Machine Intelligence},
  volume={5},
  number={12},
  pages={1344--1355},
  year={2023},
  publisher={Nature Publishing Group UK London}
}

@inproceedings{mollyn2024egotouch,
  title={{EgoTouch}: On-Body Touch Input Using {AR/VR} Headset Cameras},
  author={Mollyn, Vimal and Harrison, Chris},
  booktitle={Proceedings of the Annual ACM Symposium on User Interface Software and Technology (UIST)},
  pages={1--11},
  year={2024}
}

@inproceedings{bock2013maximum,
  title={Maximum filter vibrato suppression for onset detection},
  author={B{\"o}ck, Sebastian and Widmer, Gerhard},
  booktitle={Proceedings of the International Conference on Digital Audio Effects (DAFx)},
  volume={7},
  pages={4},
  year={2013},
  organization={Citeseer}
}

@article{carvalho2023review,
  title={Review of electromyography onset detection methods for real-time control of robotic exoskeletons},
  author={Carvalho, Camila R. and Fern{\'a}ndez, J. Marvin and Del-Ama, Antonio J. and Oliveira Barroso, Filipe and Moreno, Juan C.},
  journal={Journal of Neuroengineering and Rehabilitation},
  volume={20},
  number={1},
  pages={141},
  year={2023},
  publisher={Springer}
}

@article{hodges1996comparison,
  title={A comparison of computer-based methods for the determination of onset of muscle contraction using electromyography},
  author={Hodges, Paul W. and Bui, Bang H.},
  journal={Electroencephalography and Clinical Neurophysiology/Electromyography and Motor Control},
  volume={101},
  number={6},
  pages={511--519},
  year={1996},
  publisher={Elsevier}
}

@article{pang2022individual,
  title={Individual differences in conversational self-touch frequency correlate with state anxiety},
  author={Pang, Hio Tong and Canarslan, Feride and Chu, Mingyuan},
  journal={Journal of Nonverbal Behavior},
  volume={46},
  number={3},
  pages={299--319},
  year={2022},
  publisher={Springer}
}

@article{dreisoerner2021self,
  title={Self-soothing touch and being hugged reduce cortisol responses to stress: A randomized controlled trial on stress, physical touch, and social identity},
  author={Dreisoerner, Aljoscha and Junker, Nina M. and Schlotz, Wolff and Heimrich, Julia and Bloemeke, Svenja and Ditzen, Beate and van Dick, Rolf},
  journal={Comprehensive Psychoneuroendocrinology},
  volume={8},
  pages={100091},
  year={2021},
  publisher={Elsevier}
}

@inproceedings{mahmood2019amass,
  title={{AMASS}: Archive of motion capture as surface shapes},
  author={Mahmood, Naureen and Ghorbani, Nima and Troje, Nikolaus F. and Pons-Moll, Gerard and Black, Michael J.},
  booktitle={Proceedings of the IEEE/CVF International Conference on Computer Vision and Pattern Recognition (CVPR)},
  pages={5442--5451},
  year={2019}
}

@inproceedings{VIBE:CVPR:2020,
  title = {{VIBE}: Video Inference for Human Body Pose and Shape Estimation},
  author = {Kocabas, Muhammed and Athanasiou, Nikos and Black, Michael J.},
  booktitle = {IEEE/CVF Conference on Computer Vision and Pattern Recognition (CVPR)},
  pages = {5252--5262},
  address = {Piscataway, NJ},
  month = jun,
  year = {2020},
  doi = {10.1109/CVPR42600.2020.00530},
  month_numeric = {6}
}

@inproceedings{zhang2016skintrack,
  title={Skin{T}rack: Using the body as an electrical waveguide for continuous finger tracking on the skin},
  author={Zhang, Yang and Zhou, Junhan and Laput, Gierad and Harrison, Chris},
  booktitle={Proceedings of the ACM Conference on Human Factors in Computing Systems (CHI)},
  pages={1491--1503},
  year={2016},
  publisher={ACM},
  address={San Jose, California, USA}
}

@inproceedings{zhang2019actitouch,
  title={Acti{T}ouch: Robust touch detection for on-skin {AR/VR} interfaces},
  author={Zhang, Yang and Kienzle, Wolf and Ma, Yanjun and Ng, Shiu S and Benko, Hrvoje and Harrison, Chris},
  booktitle={Proceedings of the Annual ACM Symposium on User Interface Software and Technology (UIST)},
  pages={1151--1159},
  year={2019}
}

@article{hajika2024radarhand,
  title={{RadarHand}: A Wrist-Worn Radar for On-Skin Touch-Based Proprioceptive Gestures},
  author={Hajika, Ryo and Gunasekaran, Tamil Selvan and Haigh, Chloe Dolma Si Ying and Pai, Yun Suen and Hayashi, Eiji and Lien, Jaime and Lottridge, Danielle and Billinghurst, Mark},
  journal={ACM Transactions on Computer-Human Interaction},
  volume={31},
  number={2},
  pages={1--36},
  year={2024},
}

@inproceedings{cai2023smpler,
  title={{SMPLer-X}: Scaling up expressive human pose and shape estimation},
  author={Cai, Zhongang and Yin, Wanqi and Zeng, Ailing and Wei, Chen and Sun, Qingping and Yanjun, Wang and Pang, Hui En and Mei, Haiyi and Zhang, Mingyuan and Zhang, Lei and Loy, Chen Change and Yang, Lei and Liu, Ziwei},
  booktitle={Advances in Neural Information Processing Systems},
  volume={36},
  pages={11454--11468},
  year={2023}
}

@misc{Forte2025-Patent,
  author = {Forte, Maria-Paola and Kuchenbecker, Katherine J. and Bartels, Ulrich Jan and Ballardini, Giulia},
  title        = {System und Verfahren zur Detektion von Kontakt von Teilen eines Körpers},
  howpublished = {German Patent Application DE 10 2025 106 025.8 (provisional)},
  year         = {2025},
  month        = feb,
  note         = {Filed with the German Patent and Trademark Office on February 18, 2025, by Max-Planck-Gesellschaft zur Förderung der Wissenschaften e.V.}
}
}

\end{document}